\let\TeXyear\year
\documentclass{ieeeaccess}

\let\year\TeXyear
\usepackage{cite}
\usepackage{amsmath,amssymb,amsfonts}
\usepackage{algorithmic}
\usepackage{graphicx}
\usepackage{caption}
\usepackage{subcaption}
\usepackage{color, colortbl}
\usepackage{textcomp}
\def\BibTeX{{\rm B\kern-.05em{\sc i\kern-.025em b}\kern-.08em
  T\kern-.1667em\lower.7ex\hbox{E}\kern-.125emX}}
\usepackage{mathtools}
\usepackage[capitalize]{cleveref}
\usepackage{multirow} 
\usepackage{multicol}
\usepackage{pgfplots, pgfplotstable}
\usepackage{tikz}
\NewSpotColorSpace{PANTONE}
\AddSpotColor{PANTONE} {PANTONE3015C} {PANTONE\SpotSpace 3015\SpotSpace C} {1 0.3 0 0.2}
\SetPageColorSpace{PANTONE}
\usepackage[T1]{fontenc}

\pgfplotsset{compat=1.18}
\begin{document}

\history{Date of publication xxxx 00, 0000, date of current version xxxx 00, 0000.}
\doi{10.1109/ACCESS.2024.DOI}

\title{TrajectoryNAS: A Neural Architecture Search for Trajectory Prediction}
\author{\uppercase{Ali Asghar Sharifi}\authorrefmark{1},
\uppercase{Ali Zoljodi}\authorrefmark{1}, and Masoud Daneshtalab\authorrefmark{1}
\IEEEmembership{Senior Member, IEEE}}

\address[1]{Mälardalen University, Västerås,Sweden (e-mail: author@mdu.se)}

\markboth
{Author \headeretal: Preparation of Papers for IEEE TRANSACTIONS and JOURNALS}
{Author \headeretal: Preparation of Papers for IEEE TRANSACTIONS and JOURNALS}

\corresp{Corresponding author: Ali Zoljodi (e-mail:ali.zoljodi@mdu.se)}

\begin{abstract}
Autonomous driving systems are a rapidly evolving technology that enables driverless car production.
Trajectory prediction is a critical component of autonomous driving systems, enabling cars to anticipate the movements of surrounding objects for safe navigation. 
Trajectory prediction using Lidar point-cloud data performs better than 2D images due to providing 3D information. However, processing point-cloud data is more complicated and time-consuming than 2D images. Hence, state-of-the-art 3D trajectory predictions using point-cloud data suffer from slow and erroneous predictions.
This paper introduces TrajectoryNAS, a pioneering method that focuses on utilizing point cloud data for trajectory prediction. By leveraging Neural Architecture Search (NAS), TrajectoryNAS automates the design of trajectory prediction models, encompassing object detection, tracking, and forecasting in a cohesive manner. This approach not only addresses the complex interdependencies among these tasks but also emphasizes the importance of accuracy and efficiency in trajectory modeling. Through empirical studies, TrajectoryNAS demonstrates its effectiveness in enhancing the performance of autonomous driving systems, marking a significant advancement in the field.
Experimental results reveal that TrajcetoryNAS yield a minimum of 4.8 higger accuracy and 1.1* lower latency over competing methods on the NuScenes dataset.
\end{abstract}


\titlepgskip=-15pt

\maketitle

\section{Introduction}
\label{sec:introduction}
\PARstart{P}{redicting} future actions or states of objects around an intelligent system, such as an autonomous driving (AD) vehicle, is crucial in preventing disasters or crashes.
Driving in the real world is a stochastic process due to the presence of other vehicles and pedestrians that can take their next step resulting in accidents or congestion.
Therefore, AD systems require the crucial ability to predict the trajectory of surrounding objects \cite{liang2020pnpnet,li2020end,marchetti2020multiple}.
To perform the task of forecasting in self-driving vehicles, 2D and 3D data can be utilized. 
3D data can usually be represented in different formats, including depth images, point clouds, meshes, and volumetric grids.
The optical camera is usually good for classification tasks such as distinguishing the type of surrounding objects or detecting lane markers or traffic signs. While the performance on measuring distances and velocities is rather weak, this information can be retrieved well from radars. 
LIDARs are complementary to the other two sensors, showing competitive results.
Distances and velocities can be estimated with very high accuracy.  
Therefore, it is the preferred representation for many scene-understanding-related applications such as autonomous driving and robotics. 

Our paper presents TrajectoryNAS, an application-specific Neural Architecture Search (NAS) that aims to create a trajectory model with high accuracy and minimum displacement errors, both final and average (FDE and ADE (\cref{sec:experimental:eval})). 
Our empirical studies reveal that accurate object detection is crucial to achieve precise trajectory predictions. 
Therefore, TrajectoryNAS is designed to localize objects with a minimum error and improve the accuracy of final trajectory predictions. 
Additionally, to minimize the time required for inference, the final objective of TrajectoryNAS is to reduce the model latency.
The experimental results demonstrate that TrajcetoryNAS outperforms other approaches on the NuScenes dataset, achieving a significant increase in accuracy by at least 4.8\% and a reduction in latency by 1.1 times compared to its competitors. Finally, our contributions in this challenge can be concluded as follows:
\begin{itemize}
    \item \textbf{Novelty of TrajectoryNAS:} TrajectoryNAS, stands out as a pioneering effort in the domain of trajectory prediction for autonomous driving. Unlike previous works, our method is the first to implement Neural Architecture Search (NAS) in an end-to-end fashion, encompassing object detection, tracking, and forecasting. This comprehensive integration addresses the intricate challenges arising from the interdependencies among subtasks, such as point cloud processing, detection, and tracking, each contributing to the final trajectory forecast. Also, the complexities of the search space further amplify the complexity, making our TrajectoryNAS a unique and noteworthy contribution.
    \item \textbf{Efficient Mini Dataset Utilization:} In response to the computational demands associated with Neural architecture search, on large datasets, our method introduces an efficient two-step process. Initially, we employ a mini dataset to speed up the identification of the optimal architecture. Subsequently, the identified architecture is applied to the complete dataset, ensuring scalability and accuracy. This streamlined approach is particularly valuable when dealing with extensive datasets. revealing a noteworthy reduction in architecture search time.
    \item \textbf{Pioneering Multi-Objective Energy Function:} A key innovation of our work is the introduction of a novel multi-objective energy function. This energy function uniquely integrates considerations for object detection, tracking, forecasting, and temporal constraints. By incorporating these diverse elements into a unified framework, our approach transcends existing methodologies that often overlook the intricate relationship between these objectives. The novel energy function enhances the predictive capabilities of TrajectoryNAS, reinforcing its efficacy in real-world scenarios where temporal constraints play a crucial role.
\end{itemize}

\section{Related works}
\subsection{Trajectory Prediction}
\begin{figure}
    \centering
    \includegraphics[width=\columnwidth]{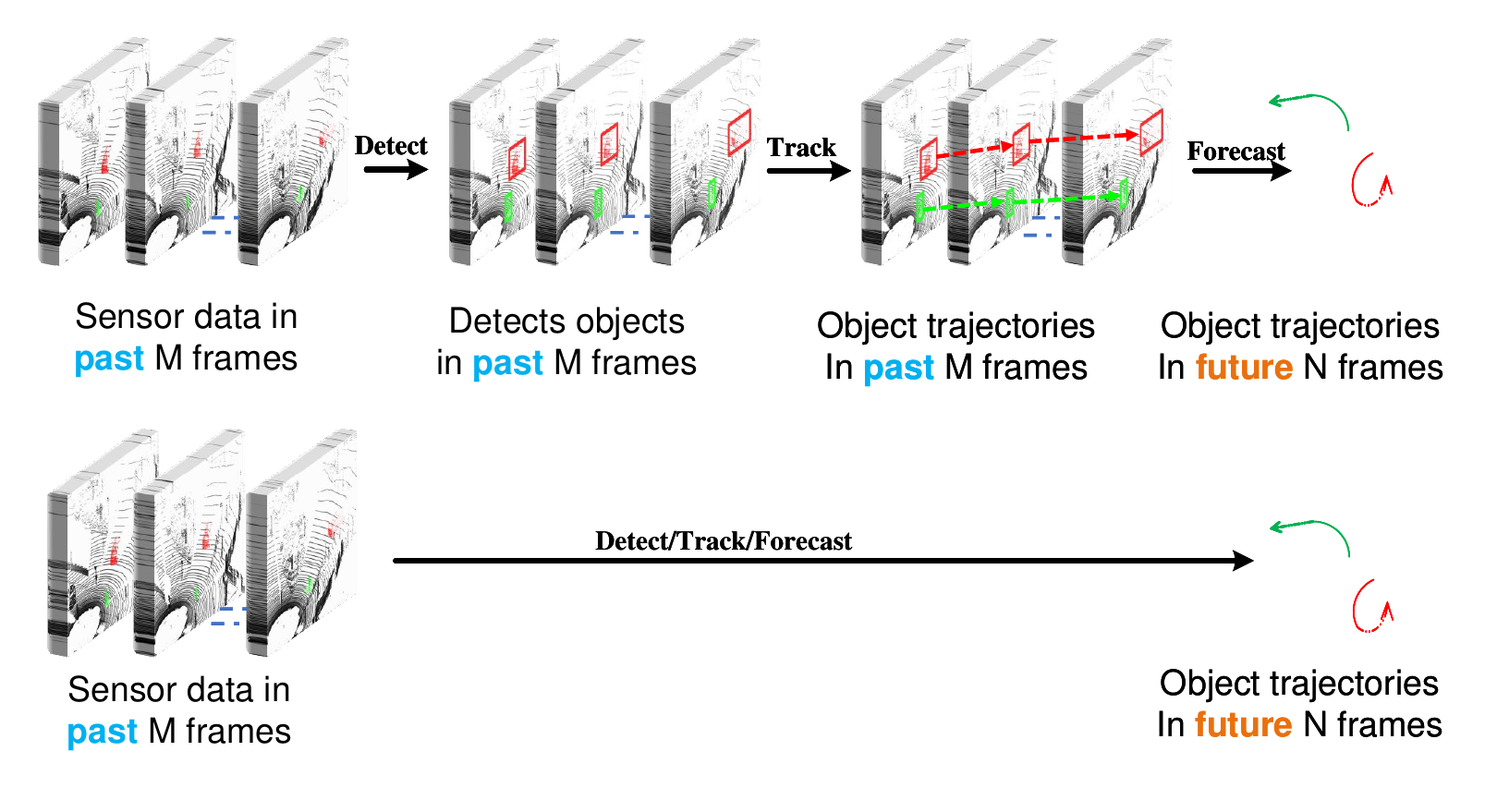}
    \caption{(Top Row) Cascade methods that independently address detection, tracking, and forecasting, they inherently carry the risk of compounding errors throughout the pipeline. This originates from each sub-module's assumption of receiving perfect input, which rarely holds true in real-world applications. Consequently, errors introduced in earlier stages propagate and magnify downstream, potentially leading to inaccurate final outcomes. (Bottom Row) End-to-end methods that forecast future movement directly from raw data, enabling end-to-end training and benefiting from the joint optimization of object detection, tracking, and prediction tasks.}
    \label{fig:Fig1}
\end{figure}

In this section, we will provide a brief overview of the literature focused on predicting trajectories using point cloud data. We begin by explaining cascade approaches (traditional approaches). In these approaches, the output of a detector serves as input to a tracker. The tracker's output is then used by a trajectory forecasting algorithm to estimate the anticipated movements of traffic participants in the upcoming seconds as in \Cref{fig:Fig1} (top row). Following that, the state-of-the-art approaches that do detection, tracking, and forecasting in an end-to-end manner are reviewed, depicted in \Cref{fig:Fig1} (bottom row) .

\begin{figure*}[ht]
\vspace{0.1cm}
    \centering
    \includegraphics[width=\textwidth]{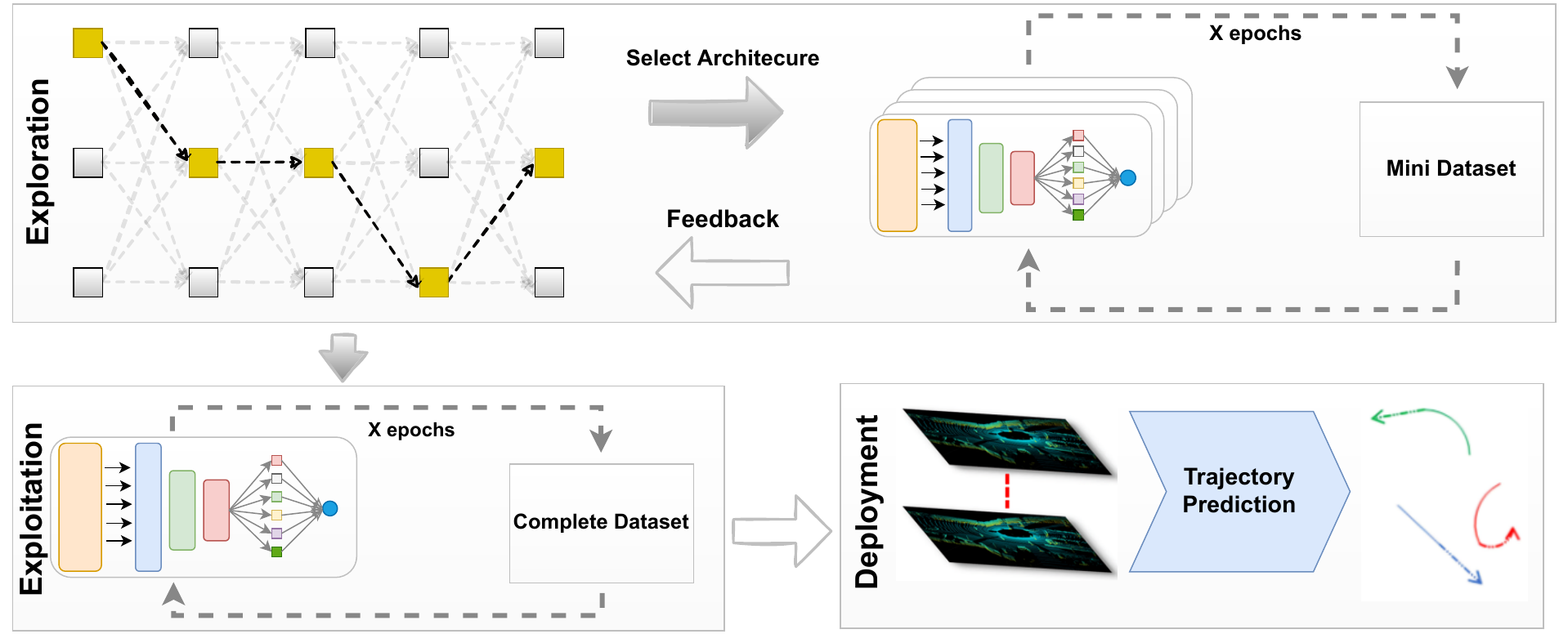}
    \caption{TrajectoryNAS state diagram. A model generated from the search space. The generated model trains using the mini dataset. The results are sent back to search space to generate a new model. The best final model is fully trained using the original dataset. }
    \label{fig:fig6}
\vspace{0.1cm}
\end{figure*}

\subsubsection{cascade approaches}
Traditional self-driving autonomy decomposes the problem into three subtasks: object detection, object tracking, and motion forecasting, and relies on independent components that perform these subtasks sequentially. These modules are usually learned independently, and uncertainty is usually propagated\cite{liang2020pnpnet}.In these methods, it is assumed that the exact paths taken by the agents are known. By examining the trajectory data over a short period of time, predictions can be made for future moments. 
For instance, the NuScenes \cite{nuscenes2019} and Argoverse \cite{Argoverse} datasets provide trajectories and their corresponding labels for this purpose.

Many of the approaches presented in the literature are based on neural networks that use recurrent neural networks (RNNs), which explicitly take into account a history composed of the past states of the agents \cite{leon2021review}. In RNNs and their variants, memory is a single hidden state vector that encodes all the temporal information. Thus, memory is addressable as a whole, and it lacks the ability to address individual elements of knowledge \cite{marchetti2020multiple}. \cite{marchetti2020multiple} presents the Memory Augmented Neural Trajectory predictor (MANTRA). In this model, an external, associative memory is trained to store useful and non-redundant trajectories. Instead of a single hidden representation addressable as a whole, the memory is element-wise addressable, permitting selective access to only relevant pieces of information at runtime.

Spatial and temporal learning will be two key components in prediction learning. Ignoring either information will lead to information loss and reduce the model's capability of context learning. Consequently, researchers are focusing on jointly learning Recurrent Neural Networks(RNN) spatial and temporal information. \cite{phan2020covernet} utilize rasterization to encode both the agents and high-definition map details, transforming corresponding elements such as lanes and crosswalks into lines and polygons of diverse colors. However, the rasterized image is an overly complex representation of environment and agent history and requires significantly more computation and data to train and deploy. In an effort to address this, VectorNet \cite{gao2020vectornet} proposes a vector representation to exploit the spatial locality of individual road components with graph neural networks. LaneConv \cite{liang2020learning} constructs a lane graph from vectorized map data and proposes LaneGCN to capture the topology and long dependency of the agents and map information. Both VectorNet \cite{gao2020vectornet} and LaneConv \cite{liang2020learning} can be viewed as extensions of graph neural networks in prediction with a strong capability to extract spatial locality. Nevertheless, both works fail to fully utilize the temporal information of agents with less focus on temporal feature extraction. In order to combine spatial and temporal learning in a flexible and unified framework, \cite{ye2021tpcn} proposes Temporal Point Cloud Networks (TPCN). TPCN models the prediction learning task as joint learning between a spatial module and a temporal module.

Across a range of visual benchmarks, transformer-based models exhibit comparable or superior performance when compared to other network types like convolutional and recurrent neural networks \cite{han2022survey}. This trend extends to trajectory prediction as well. \cite{yuan2021agentformer} proposes a new transformer that simultaneously models the time and social dimensions. Their method allows an agent's state at one time to directly affect another agent's state in the future. In parallel, \cite{khandelwal2020if} develops an RNN-based approach for context-aware multi-modal behavior forecasting. The model input includes both a road network attention module and a dynamic interaction graph to capture interpretable geometric and social relationships.

As mentioned, cascade approaches in order to trajectory prediction are developed separately from their upstream perception. As a result, their performance degrades significantly when using real-world noisy tracking results as inputs.\cite{weng2022whose} presents a novel prediction framework that uses affinity matrices rather than tracklets as inputs, thereby completely removing the chances of errors occurring in data association and passing more information to prediction. To consider this propagation of errors, \cite{weng2022whose} applies three types of data augmentation to increase the robustness of prediction with respect to tracking errors. They inject identity switches (IDS), fragments (FRAG), and noise.

\subsubsection{End-to-End approaches}
To prevent the propagation of errors and reduce inference time in traditional methods, as they learn independently, researchers
\cite{wang2020pointtracknet,yin2021center,li2023efficient,simon2019complexer} attempted to perform detection and tracking in an end-to-end manner. With the same purpose, \cite {weng2021ptp} proposed a network that parallelized tracking and prediction using a Graph Neural Network.

To our best knowledge, FaF \cite{luo2018fast} proposes the first deep neural network capable of jointly performing 3D detection, tracking, and motion forecasting using data captured by a 3D sensor. However, \cite{luo2018fast} limited its predictions to a mere 1-second duration. In contrast, IntentNet \cite{casas2018intentnet} enlarges the prediction horizon and estimates future high-level driver behavior. \cite{zeng2019end} moved a step further and performed detection, forecasting, and motion planning jointly. Furthermore,  \cite{zeng2019end} introduces an additional perception loss that encourages the intermediate representations to generate accurate 3D detections and motion forecasts. This ensures the interoperability of these intermediate representations and enables significantly accelerated learning. The statistical interconnections among actors are overlooked by all the previously mentioned methods, and instead, they individually forecast each trajectory using the provided features. \cite{li2020end} designed a novel network that explicitly takes into account the interactions among actors. To capture their spatial-temporal dependencies, \cite{li2020end} proposes a recurrent neural network with a Transformer architecture.

\cite{weng2021inverting} suggests a reversing of the detect-then-forecast pipeline rather than following the conventional sequence of detecting, tracking, and subsequently forecasting objects. Afterward, object detection and tracking are performed on the projected point cloud sequences to obtain future poses. A notable advantage of this methodology lies in the comprehensive representation of predictions, incorporating details about RNNs, the background and foreground objects existing within the scene.
Similarly, in a comparable fashion, FutureDet \cite{peri2022forecasting} directly predicts the future locations of objects observed at a specific time instead of predicting point cloud sequences over time and then backcasting them to determine their origin in the current frame. This allows the model to reason about multiple possible futures by linking future and current locations in a many-to-one manner. This approach leverages existing LiDAR detectors to predict object positions in unseen future scans. Building upon the recently proposed CenterPoint LiDAR detector\cite{Yin_2021_CVPR}, FutureDet predicts not only future locations but also velocity vectors for each object in every frame between the current and final predicted future frame. This enables the model to estimate consistent object trajectories throughout the entire forecasting horizon. In the process of forecasting, it is essential to link all trajectories to the collection of object detections in the current (observed) LiDAR scan. For each future detection i, FutureDet computes the distance to every detection j from the previous time step. Subsequently, for each i, FutureDet selects the most suitable j (permitting multiple-to-one matching).

Additionally, it is argued that current evaluation metrics for forecasting directly from raw LiDAR data are inadequate as they can be manipulated by simplistic forecasters, leading to inflated performance. These metrics, originally designed for trajectory-based forecasting, do not effectively address the interconnected tasks of detection and forecasting. To overcome these limitations, a novel evaluation procedure is proposed by FutureDet. The new metric integrates both detection and forecasting tasks. Notably, this approach surpasses state-of-the-art methods without the necessity of object tracks or HD maps as model inputs.


\subsection{Neural Architecture Search}
\label{sec:related_work:NAS}

Optimizing model hyper-parameters is an effective way to improve intelligent systems using Automated machine learning (AutoML). \cite{he2021automl}. 
Neural Architecture Search (NAS) is a subset of AutoML that aims to create efficient neural networks for complex learning tasks. \cite{elsken2019neural}. 
Early NAS methods used Reinforcement Learning (RL) \cite{zoph2016neural,hsu2018monas} or evolutionary algorithms \cite{loni2020deepmaker,loni2019neuropower}.
However, evaluating 20,000 neural architectures over four days requires remarkable computing capacity, such as 500 NVIDIA\textsuperscript{®} GPUs \cite{zoph2016neural}.
Recently, methods for differentiable neural architecture search (NAS) have been proven to achieve state-of-the-art results across various learning tasks \cite{liu2019auto,liu2018darts,loni2021TAS}. 
DARTS \cite{liu2018darts} is a differentiable NAS method that uses the gradient descent algorithm to search and train neural architecture cells jointly. 
Despite the success of differentiable NAS methods in various domains \cite{loni2021TAS}, they suffer from inefficient training due to interfering with the training of different sub-networks each other \cite{cai2019once}. 
Moreover, it has been proved that with equal search spaces and training setups, differentiable NAS methods converge to similar results \cite{NATS-bench}.

Meta-heuristic-based NAS methods~\cite{9185611,xu2020curvelane,9609009} benefit from fast and flexible algorithms to search a discrete search space. 
FastStereoNet~\cite{9609009} is a state-of-the-art meta-heuristic method that designs an accurate depth estimation pipeline. 
TrajectoryNAS is a fast multi-objective meta-heuristic NAS designed to optimize trajectory prediction approaches by searching a wider design space compared to differentiable methods or evolutionary NAS approaches.

\section{TRAJECTORYNAS}
\subsection{Framework}
Current trajectory prediction techniques rely on hand-crafted neural network architectures. These models, while effective for tasks like 3D object detection, are sub-optimal for trajectory prediction. Building on the success of Neural Architecture Search (NAS), TrajectoryNAS offers an interactive approach to designing neural networks specifically for 3D trajectory prediction. However, it's important to note that training a trajectory prediction model is both costly and time-consuming, requiring approximately 12 GPU hours for a single model. 
As a result, the NAS procedure becomes significantly slow, requiring approxmately 1200 GPU hours. To expedite the process, each model generated by the NAS is trained on a standard subset of the NuScenes~\cite{nuscenes2019} dataset. This technique reduces the evaluation time for each model to nearly 1 hour, making the process approximately 12 times faster.

\Cref{fig:fig6} elaborates TrajectoryNAS state diagram.

TrajectoryNAS is a one-stage trajectory prediction \cite{peri2022forecasting,luo2018fast}. The model takes a sequence of Lidar data captured from the scene, which integrates a robust 3D backbone with cutting-edge Neural Architecture Search (NAS) to refine map-view feature extraction from LiDAR point clouds. This innovative architecture further evolves by automating the design of multi-2D CNN detection heads, specifically tailored for future object detection and trajectory prediction. By detecting objects across multiple future timesteps and accurately projecting their movements back to the current moment, TrajectoryNAS stands out for its precision in trajectory forecasting. This system not only anticipates the dynamic positioning of objects but also adjusts its computational strategies in real-time, ensuring a high degree of accuracy and efficiency in processing. The inclusion of NAS allows for continuous improvement of the detection and forecasting heads, making TrajectoryNAS a highly adaptive and forward-thinking solution in the realm of autonomous navigation and surveillance technologies.

\begin{figure*}[ht]
\vspace{0.1cm}
    \centering
    \includegraphics[width=\textwidth]{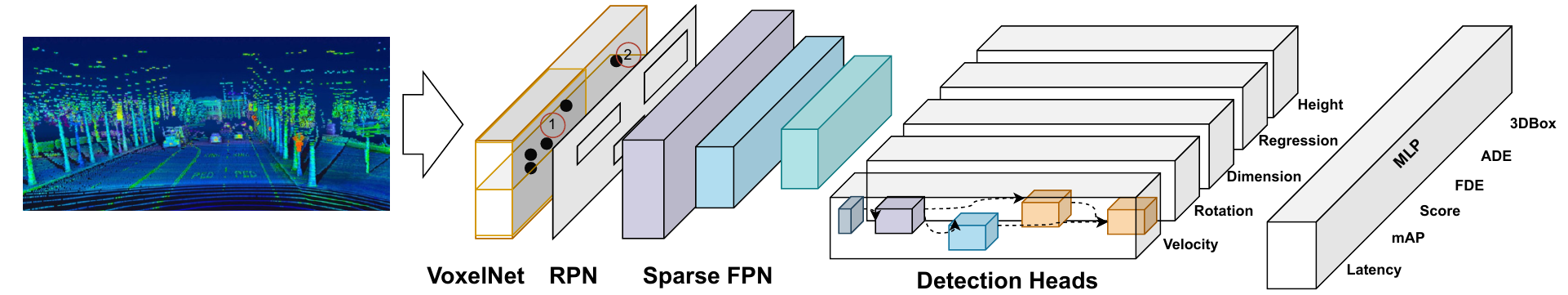}
    \caption{The overview of TrajcetoryNAS process.}
    \label{fig:fig6_1}
\vspace{0.1cm}
\end{figure*}

\subsection{Search Space}


The TrajectryNAS architecture stands out as a solution for object detection and trajectory prediction, particularly in scenarios like autonomous driving where understanding dynamic environments is paramount. It skillfully merges spatial and temporal object analyses, predicting not only the present state but also future trajectories. At its core, the VoxelNet Backbone transforms point clouds into structured voxel representations, enabling the extraction of comprehensive spatial features. This is complemented by a Sparse Feature Pyramid Network (FPN), which builds a pyramid of varying resolution features while efficiently managing computational complexity by concentrating on relevant areas within the voxel space.

TrajectoryNAS contains five parallel prediction heads, each dedicated to a specific aspect of object state: Velocity, Rotation, Dimension, Regression, and Height. These heads work in tandem to provide a multifaceted description of an object's current position and orientation at time $t$. The architecture's prowess extends to predicting future object trajectories. By leveraging the initial predictions, it projects the object's state to time $t+1$, then cyclically feeds this data back into the prediction heads to forecast the object's path over future steps.

Such an approach allows TrajectryNAS to not only navigate but also anticipate complex dynamic behaviors, making it an invaluable asset in fields where predicting future states is crucial for proactive decision-making. This architecture's ability to foresee the direction and movement of objects enriches scene understanding and enhances planning for autonomous systems, offering a comprehensive and forward-looking perspective on environmental dynamics.

The TrajectoryNAS system automatically designs the Region Proposal Network (RPN) and the prediction heads using the aforementioned layers. It explores an expansive space of $2^{300}$ potential architectures to identify an optimal balance between speed and accuracy. This approach enables the selection of a highly efficient and accurate architecture tailored for specific applications.

\subsection{Search Algorithm}

To improve the accuracy of trajectory prediction while reducing network inference time, we employ the multi-objective simulated annealing (MOSA) algorithm, as described in \cite{amine2019multiobjective}. MOSA selects candidates based on the probability of $min(1, exp(-\Delta/T))$, where $\Delta$ is the energy difference between present and newly generated candidates, and $T$ is the regulating parameter for annealing temperature. Initially, $T$ starts from a large value ($T_{Max}$) and gradually decreases to a small value ($T_{Min}$). Setting $T_{Max}$ to a large value allows for exploration of non-optimal choices, while $T_{Min}$ being small gives the maximum selection chance to optimal candidates (exploitation).

To achieve this optimization, we use a multi-objective energy function (Eq.~\ref{sec:Energy:eq2}). 
The fitness function ($F$) is the product of the network latency ($t$) and the weighted mean average precision of the predicted future place of the object and its actual place (mAP), weighted average displacement (ADE) error, and weighted final displacement error (FDE). 
\begin{align}
\label{sec:Energy:eq2}
\mathcal{F}={Latency}\times{mAP}^\alpha\times{ADE}^\beta\times{FDE}^\gamma
\end{align}
Where $\alpha$, $\beta$, and $\gamma$ are weights of mAP, ADE, and FDE respectively.
We do not use any proxy, such as \textit{Floating-Point-Operations-per-Second} (FLOPs), for inference time estimation. Instead, we run the network directly on the target hardware (NVIDIA\textsuperscript{®} RTX A4000) to measure the exact inference time.

\section{EXPERIMENTAL SETUP}
\label{sec:experimental}
We demonstrate the effectiveness of our approach on a large-scale real-world driving dataset. We focus on modular metrics for detection and forecasting, as well as system metrics for end-to-end perception and prediction.
\subsection{Dataset}
\label{sec:experimental:dataset}
Our experimental analysis was performed on the nuScenes \cite{nuscenes2019} dataset, which contains 1000 log snippets, each lasting 20 seconds. We utilized two officially released divisions of the dataset: the Mini and Trainval splits. The Mini split, which consists of 10 scenes, is a subset of the Trainval split. The Trainval split contains 700 scenes for training purposes and 150 scenes for validation. Additionally, the Test split, containing 150 scenes, is designated for challenges and lacks object annotations.
\subsection{Evaluation Metrics}
\label{sec:experimental:eval}
The performance of the End-to-End object forecasting method is assessed using two standard metrics: Average Precision for object detection ({${AP}_{det}$}) and Forecasting Average Precision ({${AP}_{f}$}). Inspired by FuterDet \cite{peri2022forecasting}, we have defined three subclasses: static cars, linearly moving cars, and non-linearly moving cars, and we report {${AP}_{f}$} and {${AP}_{det}$} for these three classes. Subsequently, we evaluate the mean Average Precision for forecasting ({${mAP}_{f}$}) as follows {${mAP}_{f} =1/3 \times ({AP}_{f}^{lin.} + {AP}_{f}^{non-lin.} + {AP}_{f}^{stat.} )$}. Similarly, {${mAP}_{det}$} is evaluated as the average {${AP}_{det}$} over the three subclasses.
\subsection{Configuration Setup}
For this study, \cref{tab:1} provides a brief overview of the configuration setup.

\begin{table}[ht]
  \caption{Summarizing hardware specification, train, and search parameters.}
  \label{tab:1}
  \begin{center}
  \begin{tabular}{c c}
    \hline
    \textbf{Train/Test Hardware Device} & \textbf{Specification}\\
    \hline
   GPU & NVIDIA\textsuperscript{®} RTX A4000 \\
   GPU Compiler & CUDA v11.7 \& cuDNN v8.2.0\\
   DL Framework & PyTorch v1.9.1\\
   
  \hline
   \textbf{Training and Search Parameters} & \textbf{Value}\\
    \hline
    Full-Training Epochs & 20\\
    Batch Size & 1 \\
    Learning Rate & $5\times10^{-4}$\\
    Optimizer & Adam\\
    $T_{Max}$ / $T_{Min}$  & 2500 / 2.5\\
  \hline
    \end{tabular}  
  \end{center}
  \vspace{-0.5cm}
\end{table}

\section{RESULTS}
\subsection{Trajectory Prediction Performance}
As presented in \cref{tab:4,tab:5}, the comparison of car and pedestrian trajectory prediction results demonstrates that TrajectoryNAS outperforms other state-of-the-art trajectory prediction methods in numerous parameters for car trajectory prediction and the majority of parameters for pedestrian trajectory prediction. Notably, the latency of TrajectoryNAS is comparable to that of Fast and Furious \cite{luo2018fast} and better than FutureDet \cite{peri2022forecasting}, while TrajectoryNAS provides superior future average precision (\(AP_f\)) across all conditions for both linear and non-linear trajectories of cars and pedestrians.

For cars, while Fast and Furious and FutureDet offer a marginal improvement in specific aspects when compared with TrajectoryNAS, TrajectoryNAS significantly surpasses the state-of-the-art in most parameters. This is evidenced by its top performance in average precision for static, linear, and non-linear trajectories, as well as its mean average precision (mAP), both for single (\(K=1\)) and multiple (\(K=5\)) predictions. Specifically, TrajectoryNAS achieves the highest detection accuracy and future average precision in almost all scenarios, highlighting its robustness and efficacy in car trajectory prediction.

Similarly, for pedestrian trajectory prediction, TrajectoryNAS demonstrates outstanding performance, particularly in accurately predicting linear and non-linear movements. It not only achieves the highest average precision scores across various scenarios but also maintains a competitive latency, underscoring its effectiveness in real-time applications.

In conclusion, TrajectoryNAS advances the field of trajectory prediction by offering a highly accurate and efficient model. Its ability to provide better future average precision under different conditions for both cars and pedestrians, coupled with its comparable latency to leading models, positions TrajectoryNAS as a superior choice for trajectory prediction in dynamic environments.
\begin{table*}[t]
\vspace{0.7cm}
  \caption{Comparison TrajcetoryNAS and state-of-the-art trajectory prediction model on cars according to accuracy and latency metrics.}
  \label{tab:4}
  \renewcommand{\arraystretch}{1.8}
  \centering
 \resizebox{\textwidth}{!}{%
  \begin{tabular}{c|c| c|c|c| c |c |c| c| c |c| c |c| c| c|c| c|c }
 
  \hline
  \multirow{3}{*}{\rotatebox{35}{\textbf{Method}}}&\multirow{3}{*}{\rotatebox{35}{\textbf{Time (ms)}}}&\multicolumn{8}{c|}{$K=1$}&\multicolumn{8}{c}{$K=5$}\\
  
\cline {3-18}

  &&\multicolumn{2}{c|}{${AP}^{stat.}$}&\multicolumn{2}{c|}{${AP}^{lin.}$}&\multicolumn{2}{c|}{${AP}^{non-lin.}$}&\multicolumn{2}{c|}{$mAP$}&\multicolumn{2}{c|}{${AP}^{stat.}$}&\multicolumn{2}{c|}{${AP}^{lin.}$}&\multicolumn{2}{c|}{${AP}^{non-lin.}$}&\multicolumn{2}{c}{$mAP$}\\
\cline {3-18}  &&${AP}_{det.}$&${AP}_{f}$&${AP}_{det.}$&${AP}_{f}$&${AP}_{det.}$&${AP}_{f}$&${AP}_{det.}$&${AP}_{f}$&${AP}_{det.}$&${AP}_{f}$&${AP}_{det.}$&${AP}_{f}$&${AP}_{det.}$&${AP}_{f}$&${AP}_{det.}$&${AP}_{f}$\\
\hline
Detection + Constant Velocity &\cellcolor{cyan!25}21&70.3&\cellcolor{cyan!25}66.0&65.8&21.2&90.0&6.5&\cellcolor{cyan!25}75.4&31.12&70.3&66.0&65.8&21.2&90.0&6.5&\cellcolor{cyan!25}75.4&31.2\\
\hline
Detection + Forecast ~\cite{luo2018fast} &20&69.1& 64.7&\cellcolor{cyan!25} 66.1& 22.2& 86.3& 7.5& 73.8& 31.5& 69.1 &64.7& \cellcolor{cyan!25}66.1& 22.2& 86.3& 7.5& 73.8& 31.5\\
\hline               
FutureDet~\cite{peri2022forecasting}&24&70.0&65.5&62.9&24.9&\cellcolor{cyan!25}91.8&10.1&74.9&33.5&70.1&67.3&62.9&27.7&\cellcolor{cyan!25}91.7&11.7&74.9&35.6\\
\hline  
TrajectoryNAS (ours)&\textbf{22}&\cellcolor{cyan!25}\textbf{71.0}&\textbf{65.6}&\textbf{63.8}&\cellcolor{cyan!25}\textbf{26}&\textbf{91.2}&\cellcolor{cyan!25}\textbf{10.3}&\textbf{75}&\cellcolor{cyan!25}\textbf{34}&\cellcolor{cyan!25}\textbf{71}&\cellcolor{cyan!25}\textbf{67.4}&\textbf{63.8}&\cellcolor{cyan!25}\textbf{29.2}&\textbf{91.1}&\cellcolor{cyan!25}\textbf{12.1}&\textbf{75.3}&\cellcolor{cyan!25}\textbf{36.2}\\
   \hline
\end{tabular}
}
\vspace{0.9cm}
\end{table*}

\begin{table*}[t]
\vspace{0.7cm}
  \caption{Comparison TrajcetoryNAS and state-of-the-art trajectory prediction model on pedestrian according to accuracy and latency metrics.}
  \label{tab:5}
  \renewcommand{\arraystretch}{1.8}
  \centering
 \resizebox{\textwidth}{!}{%
  \begin{tabular}{c|c| c|c|c| c |c |c| c| c |c| c |c| c| c|c| c|c }
 
  \hline
  \multirow{3}{*}{\rotatebox{35}{\textbf{Method}}}&\multirow{3}{*}{\rotatebox{35}{\textbf{Time (ms)}}}&\multicolumn{8}{c|}{$K=1$}&\multicolumn{8}{c}{$K=5$}\\
  
\cline {3-18}

  &&\multicolumn{2}{c|}{${AP}^{stat.}$}&\multicolumn{2}{c|}{${AP}^{lin.}$}&\multicolumn{2}{c|}{${AP}^{non-lin.}$}&\multicolumn{2}{c|}{$mAP$}&\multicolumn{2}{c|}{${AP}^{stat.}$}&\multicolumn{2}{c|}{${AP}^{lin.}$}&\multicolumn{2}{c|}{${AP}^{non-lin.}$}&\multicolumn{2}{c}{$mAP$}\\
\cline {3-18}  &&${AP}_{det.}$&${AP}_{f}$&${AP}_{det.}$&${AP}_{f}$&${AP}_{det.}$&${AP}_{f}$&${AP}_{det.}$&${AP}_{f}$&${AP}_{det.}$&${AP}_{f}$&${AP}_{det.}$&${AP}_{f}$&${AP}_{det.}$&${AP}_{f}$&${AP}_{det.}$&${AP}_{f}$\\
\hline
Detection + Constant Velocity &\cellcolor{cyan!25}21&55.1&33.3&73.5&27.8&96.9&12.4&75.2&25.5&55.1&33.3&73.5&27.8&96.9&12.4&75.2&24.5\\
\hline
Detection + Forecast ~\cite{luo2018fast} &20&53.7& 35.0& 73.9&30.8&\cellcolor{cyan!25} 97.2& 13.3& 74.9& 26.4& 53.7 &35.0& 73.9& 30.8& \cellcolor{cyan!25}97.2& 13.3& 74.9& 26.4\\
\hline               
FutureDet~\cite{peri2022forecasting}&24&53.1&33.3&72.4&32.6&95.2&14.7&73.6&26.9&53.1&35.1&72.4&34.0&95.2&15.0&73.6&28.0\\
\hline  
TrajectoryNAS (ours)&\textbf{22}&\cellcolor{cyan!25}\textbf{55.8}&\cellcolor{cyan!25}\textbf{37.1}&\cellcolor{cyan!25}\textbf{77.9}&\cellcolor{cyan!25}\textbf{39.9}&\textbf{95.2}&\cellcolor{cyan!25}\textbf{17.7}&\cellcolor{cyan!25}\textbf{76.3}&\cellcolor{cyan!25}\textbf{31.3}&\cellcolor{cyan!25}\textbf{55.8}&\cellcolor{cyan!25}\textbf{38.6}&\cellcolor{cyan!25}\textbf{77.9}&\cellcolor{cyan!25}\textbf{40.9}&\textbf{95.2}&\cellcolor{cyan!25}\textbf{17.9}&\cellcolor{cyan!25}\textbf{76.3}&\cellcolor{cyan!25}\textbf{32.5}\\
   \hline
\end{tabular}
}
\vspace{0.9cm}
\end{table*}
\subsection{Analysing Search Methods}
\begin{figure*}[ht]
    \begin{tikzpicture}
      \begin{axis}[
        enlarge x limits=false,
        column sep=4em,
        height=7cm,
        width=\textwidth, 
        ymin=0.10, ymax=0.33,
        xmin=-3,xmax=110,
        grid=both, 
        minor grid style={dotted,gray!60}, 
        major grid style={dashed,gray!70}, 
        minor tick num=4, 
        xlabel= \textbf{Search iterations},
        ylabel= \textbf{Energy function},
        tick label style={font=\large},
        ylabel near ticks,
        xlabel near ticks,
        legend style={anchor=north,legend columns=3},
        legend cell align={left},
        legend style={draw=black, at={(0.5,1.0)}, text opacity = 1,row sep=0pt, font=\fontsize{9}{6}\selectfont},
         x tick label style={rotate=0,anchor=north},
        every axis x label/.style={at={(0.5,-0.15)},anchor=south}
        ]
        \addplot [ mark size=4pt, red, thick, draw=red,opacity=0.2 ] coordinates {
(0,0.2900) (1,0.1518) (2,0.1420) (3,0.1166) (4,0.1400) (5,0.1339) (6,0.1662) (7,0.1711) (8,0.1646) (9,0.1784) (10,0.1583) (11,0.2717) (12,0.1745) (13,0.1710) (14,0.1378) (15,0.1389) (16,0.1341) (17,0.2557) (18,0.1377) (19,0.1749) (20,0.1526) (21,0.1491) (22,0.1514) (23,0.1480) (24,0.1612) (25,0.1318) (26,0.1509) (27,0.1394) (28,0.1737) (29,0.1781) (30,0.1454) (31,0.1509) (32,0.1604) (33,0.1141) (34,0.1343) (35,0.1448) (36,0.1387) (37,0.1628) (38,0.1636) (39,0.1645) (40,0.1423) (41,0.1582) (42,0.1422) (43,0.1432) (44,0.1538) (45,0.1602) (46,0.1684) (47,0.1533) (48,0.1774) (49,0.1462) (50,0.1487) (51,0.1495) (52,0.1714) (53,0.1726) (54,0.1789) (55,0.1763) (56,0.1344) (57,0.1573) (58,0.1691) (59,0.1549) (60,0.1316) (61,0.1522) (62,0.1423) (63,0.1484) (64,0.1532) (65,0.1395) (66,0.1790) (67,0.1311) (68,0.1727) (69,0.1385) (70,0.1673) (71,0.1506) (72,0.1568) (73,0.1572) (74,0.1584) (75,0.1484) (76,0.1317) (77,0.1500) (78,0.1314) (79,0.1581) (80,0.1484) (81,0.1543) (82,0.1504) (83,0.1558) (84,0.1582) (85,0.1428) (86,0.1485) (87,0.1583) (88,0.1448) (89,0.1577) (90,0.1420) (91,0.1385) (92,0.1608) (93,0.1350) (94,0.1602) (95,0.1400) (96,0.1346) (97,0.1498) (98,0.1309) (99,0.1388) (100,0.1327) (101,0.1548) (102,0.1345) (103,0.1309) (104,0.1552) (105,0.1330) (106,0.1346) (107,0.1307) (108,0.1130)
};
        \addplot [mark=square*, mark size=4pt, red, thick , draw=red ] coordinates {(0,0.29) (1,0.1518)(2,0.1420)(3,0.1166)(33,0.1141)(108,0.113)};
        \addplot [ mark size=4pt, red, thick , draw=blue,opacity=0.2 ] coordinates {
(0,0.2900) (1,0.23) (2,0.2420) (3,0.2566) (4,0.2400) (5,0.2679) (6,0.2681) (7,0.2544) (8,0.2517) (9,0.22) (10,0.2724) (11,0.2747) (12,0.2765) (13,0.2605) (14,0.2601) (15,0.2719) (16,0.2660) (17,0.2681) (18,0.2718) (19,0.2756) (20,0.2535) (21,0.2550) (22,0.2438) (23,0.2631) (24,0.2414) (25,0.2586) (26,0.2617) (27,0.2515) (28,0.20) (29,0.2412) (30,0.2415) (31,0.2729) (32,0.2544) (33,0.2141) (34,0.2451) (35,0.2609) (36,0.2708) (37,0.2486) (38,0.2649) (39,0.2434) (40,0.2421) (41,0.2613) (42,0.2616) (43,0.2655) (44,0.2690) (45,0.2790) (46,0.2607) (47,0.2529) (48,0.2718) (49,0.2508) (50,0.2576) (51,0.2431) (52,0.2410) (53,0.19) (54,0.2734) (55,0.2678) (56,0.2564) (57,0.2469) (58,0.2463) (59,0.2500) (60,0.2620) (61,0.2686) (62,0.2664) (63,0.2512) (64,0.2782) (65,0.2695) (66,0.2622) (67,0.2645) (68,0.2568) (69,0.2499) (70,0.2542) (71,0.2703) (72,0.2406) (73,0.2446) (74,0.2418) (75,0.2416) (76,0.2742) (77,0.2681) (78,0.2590) (79,0.2439) (80,0.2597) (81,0.2589) (82,0.2469) (83,0.2574) (84,0.2559) (85,0.2646) (86,0.2654) (87,0.2418) (88,0.2550) (89,0.2650) (90,0.2601) (91,0.2743) (92,0.2663) (93,0.2465) (94,0.2428) (95,0.2657) (96,0.2411) (97,0.2634) (98,0.2776) (99,0.2630) (100,0.2555) (101,0.2657) (102,0.2583) (103,0.2618) (104,0.2777) (105,0.2554) (106,0.2784) (107,0.2762) (108,0.2730)
};

        \addplot[mark=diamond*, mark size=4pt, blue, thick , draw=blue ] coordinates {(0,0.29)
        (1,0.23)(9,0.22)(28,0.20)(53,0.19)};
        \addplot[mark=triangle*, mark size=4pt, orange, thick , draw=orange ] coordinates {(0,0.29)
        (1,0.26)(11,0.24)(33,0.21)(45,0.19)(50,0.186)};
        
\addplot[mark size=4pt, orange, thick , draw=orange ,opacity=0.2 ] coordinates {
(0,0.2900) (1,0.2600) (2,0.2478) (3,0.2428) (4,0.2440) (5,0.2407) (6,0.2438) (7,0.2673) (8,0.2428) (9,0.2528) (10,0.2738) (11,0.2400) (12,0.2409) (13,0.2726) (14,0.2513) (15,0.2447) (16,0.2679) (17,0.2652) (18,0.2751) (19,0.2694) (20,0.2721) (21,0.2513) (22,0.2471) (23,0.2700) (24,0.2723) (25,0.2796) (26,0.2565) (27,0.2549) (28,0.2711) (29,0.2536) (30,0.2772) (31,0.2743) (32,0.2572) (33,0.2100) (34,0.2700) (35,0.2702) (36,0.2441) (37,0.2761) (38,0.2602) (39,0.2731) (40,0.2528) (41,0.2758) (42,0.2556) (43,0.2404) (44,0.2762) (45,0.1900) (46,0.2437) (47,0.2528) (48,0.2780) (49,0.2780) (50,0.1860) (51,0.2629) (52,0.2653) (53,0.2579) (54,0.2517) (55,0.2531) (56,0.2669) (57,0.2701) (58,0.2717) (59,0.2716) (60,0.2436) (61,0.2598) (62,0.2423) (63,0.2620) (64,0.2577) (65,0.2755) (66,0.2540) (67,0.2447) (68,0.2457) (69,0.2405) (70,0.2647) (71,0.2440) (72,0.2434) (73,0.2680) (74,0.2429) (75,0.2729) (76,0.2682) (77,0.2433) (78,0.2434) (79,0.2795) (80,0.2550) (81,0.2548) (82,0.2725) (83,0.2779) (84,0.2794) (85,0.2701) (86,0.2551) (87,0.2433) (88,0.2211) (89,0.2023) (90,0.2570) (91,0.2763) (92,0.2444) (93,0.2597) (94,0.2405) (95,0.2587) (96,0.2423) (97,0.2448) (98,0.2447) (99,0.2660) (100,0.2698) (101,0.2633) (102,0.2785) (103,0.2550) (104,0.2514) (105,0.2747) (106,0.2489) (107,0.2785) (108,0.2405) (109,0.2788) (110,0.2417)
};
\legend{,TrajectoryNAS,,Random Search ,Local Search}

      \end{axis}
    \end{tikzpicture}  
    \captionof{figure}{TrajectoryNAS optimization curve.}
    \label{fig:discussion:reproducibility}
\end{figure*}
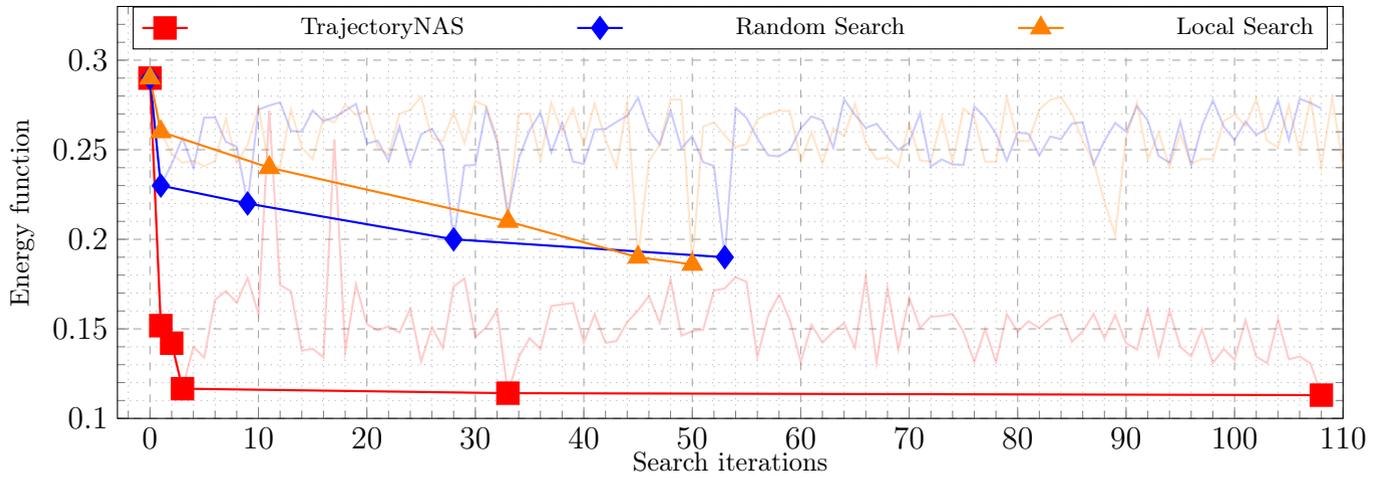

Figure \ref{fig:discussion:reproducibility} presents a detailed comparison of the energy function reduction (as defined in Eq.~\ref{sec:Energy:eq2}) during the search process employed by the TrajectoryNAS algorithm against those of Random Search and Local Search methods. This comparative analysis clearly demonstrates the limitations of both Local Search and Random Search techniques in effectively identifying the most optimal solution. Specifically, the best outcome identified through Random Search, characterized by an energy value of e=0.19 as per \cref{sec:Energy:eq2}, was achieved in iteration 52. Similarly, Local Search's most effective solution registered an energy value of e=0.186, and this result was obtained in iteration 50.

Despite these efforts, both methods fall significantly short when compared to the capabilities of the TrajectoryNAS algorithm. TrajectoryNAS not only surpasses these traditional search methodologies in efficiently navigating towards more optimal solutions but also showcases its superiority by discovering an exceptionally lower energy value of 0.113. This landmark achievement was realized in iteration 108, underlining the algorithm's advanced optimization prowess. Notably, the energy value associated with the best solution found by TrajectoryNAS is nearly half that of the best solutions unearthed by both Random Search and Local Search. This stark contrast underscores the advanced and sophisticated nature of TrajectoryNAS in exploring and exploiting the search space to find significantly more efficient solutions, thereby establishing a new benchmark in the quest for optimization within this context.

\section{CONCLUSION}
Our paper presents TrajectoryNAS, an automated model design approach that significantly enhances 3D trajectory prediction for autonomous driving. By optimizing for both speed and accuracy while considering key performance metrics, TrajectoryNAS outperforms existing methods by achieving a minimum of 4.8\% higher accuracy and 1.1 times lower latency on the NuScenes dataset.

\newpage
\bibliographystyle{IEEEtran}  
\bibliography{references.bib}

\begin{IEEEbiography}[{\includegraphics[width=1in,height=1.25in,clip,keepaspectratio]{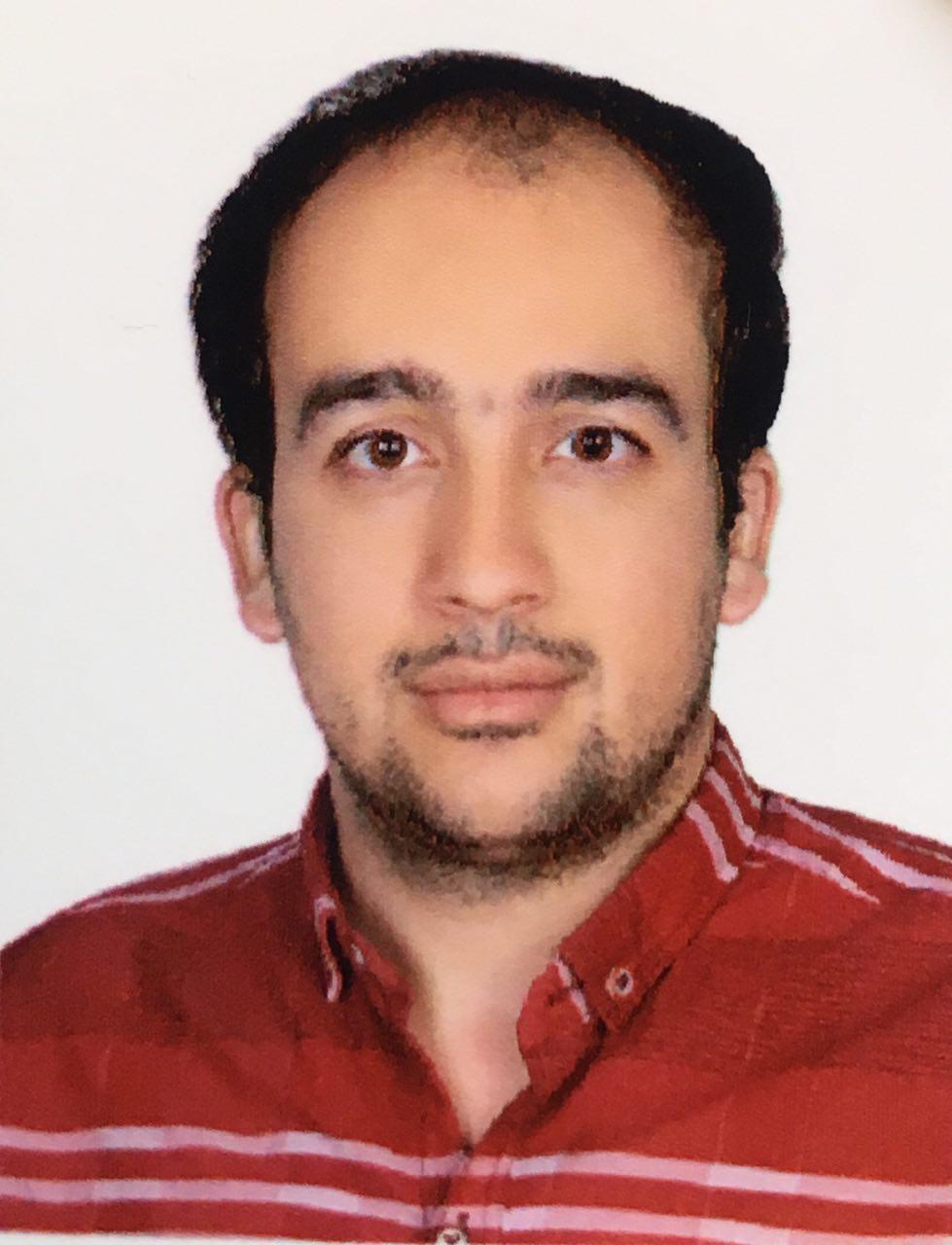}}]{Ali Asghar Sharifi}  received his B.Sc. and M.Sc. degrees in Communications Engineering from Arak University, Arak, Iran, in 2015, and Communication Engineering with a specialization in Computer Vision from Shahid Beheshti University, Tehran, Iran, in 2017, respectively.
He joined the Heterogeneous System Research Group as an ML Engineer in Västerås, Sweden, in 2021, focusing on trajectory prediction and object detection for autonomous driving systems. Prior to this, he served as a Research Assistant at the Digital Signal Processing Laboratory (DiSPLaY) Group in Tehran, Iran, from 2017 to 2021, where his work involved various aspects of digital signal and image processing. His research interests span machine learning, deep learning, computer vision, and signal processing, as evidenced by his contributions to prestigious IEEE publications.
\end{IEEEbiography}

\begin{IEEEbiography}[{\includegraphics[width=1in,height=1.25in,clip,keepaspectratio]{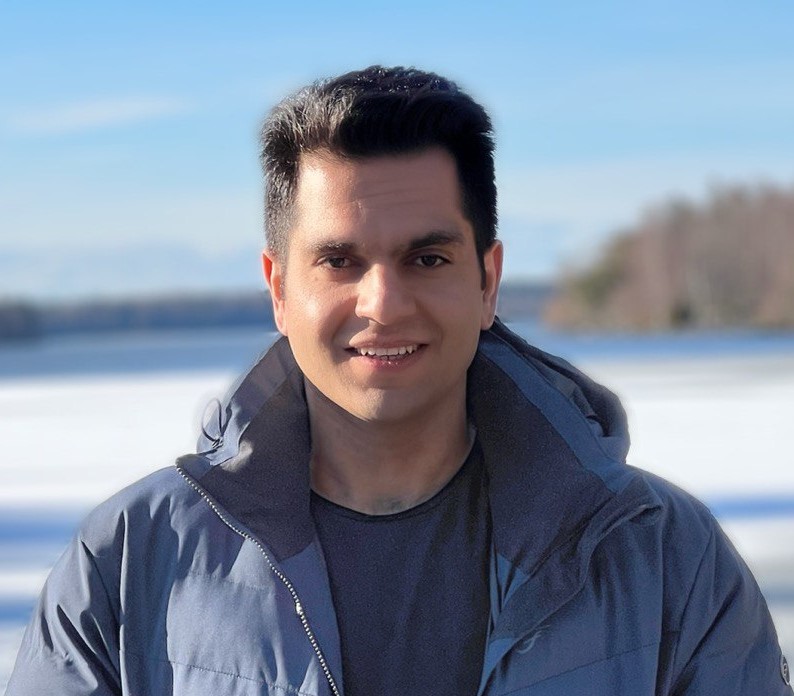}}]{Ali Zoljodi } is Ph.D. student in Artificial Intelligence at Mälardalen University, Västerås, Sweden. His main research interest is Autonomous driving systems. He is currently working on lane detection perception. During his Ph.D. studies, he published a paper on  3D lane detection enhancement using Neural Architecture Search (NAS). He also has a ready-to-submit paper on using self-supervised learning in lane detection applications. Before his Ph.D. studies, Ali Zoljodi experienced developing NAS for various applications such as image classification and Stereo Vision algorithms.
\end{IEEEbiography}

\begin{IEEEbiography}[{\includegraphics[width=1in,height=1.25in,clip,keepaspectratio]{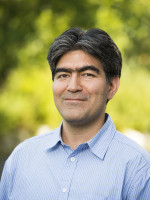}}]{Masoud Daneshtalab} received the Ph.D. degree in computer science from the University of Turku, Turku, Finland, in 2011. He is currently a Professor with Mälardalen University (MDU), Västerås, Sweden. He is an
Adjacent Professor with TalTech, Tallinn, Estonia, and co-leading the Heterogeneous System Research Group. He is on the Euromicro board of directors and has published over 200 refereed papers. His research interests include HW/SW co-design, reliability, and deep learning acceleration. Prof. Daneshtalab is an Editor of the Microprocessors and Microsystems: Embedded Hardware Design.
\end{IEEEbiography}

\EOD

\end{document}